# Automatic Sentence Completion Using Bengali Audio Speech with a Transformer-Based Early Speech Recognition System

Md Asif Akram[1], Arif Faisal[1], Sakir Hossain Faruque[1], Sharun Akter Khushbu[1]

[1]Department of Computer Science and Engineering, Daffodil International University, Dhaka-1216, Bangladesh,

**Abstract**
Dementia severely impairs communication abilities and turns everyday conversations into challenging experiences. Individuals affected by dementia frequently struggle to express their thoughts or complete sentences, making it difficult for family members and caregivers to understand their needs. This study proposes a novel approach to assist dementia patients by developing an integrated system that detects signs of dementia through Bangla speech and text analysis and provides sentence auto-completion support during conversations. To achieve this, recorded voices from individuals with and without dementia were analysed, and key acoustic features were extracted to capture vocal patterns associated with cognitive decline. A transformer-based model was employed to transcribe voice signals into text for linguistic analysis. In the classification task, several Machine Learning (ML) and Deep Learning (DL) models were evaluated using both acoustic and textual representations, and the LSTM model architectures demonstrated strong capabilities for modelling contextual patterns in speech and language data. For the sentence auto-completion task, a sequential learning model based on a Bidirectional LSTM (Bi-LSTM) network with an attention mechanism was developed to sequentially complete partially expressed Bangla sentences during conversation. This research bridges the gap in dementia analysis in an underrepresented language and implements a sentence auto-completion framework for cognitive disorders.

## 1- Introduction

Dementia is an age-related brain condition that interferes with a person's ability to think, remember and reason. It's not a single disease — it is a set of symptoms that can manifest in multiple ways. The World Health Organization says depression is becoming a global problem. It is diagnosed every three seconds (Alzheimer's Disease International, n.d.). When brain cells no longer speak to one another as they should, dementia takes hold, and it becomes more difficult for people to look after themselves day in and day out.

Worldwide, dementia is becoming more common, just like in Bangladesh. Here, one in 12 adults aged 60 or older has dementia (The Business Standard, 2021). This condition is often mistaken for ordinary age-related changes, yet it stems from serious brain illnesses (Harvard Health Publishing, 2022). By 2022, China had 15.3 million older adults - about 9.28% of its overall population - a number expected to climb toward 18% by mid-century, adding pressure (The Daily Star, 2023). Worldwide, dementia numbers jumped from 43.8 million in 2016 to 57.6 million just three years later, adding close to 10 million annually (Collaborators G. 2., 2019; 2022), while forecasts suggest they could reach 132 million by the century mark (World Health Organization, 2017). That year, 2020, dementia led to 14,993 fatalities in Bangladesh - making up 2.09% of all deaths - while in the U.S., it caused 287,198 such losses (World Life Expectancy, 2020), showing just how heavy the toll has grown.

Living with dementia makes communication harder with other people over time. When the illness moves forward, ordinary chats turn tricky - words fade, moments stretch, ideas get tangled. Struggling to speak brings tension, isolation, and silence where connection once stood (World Health Organization, 2025). What makes things more complicated in Bangladesh is how often people think dementia signs are only old age wearing thin. That false assumption leads families to wait too long before getting help, ending up with diagnoses too late and little support available - even more so where knowledge and services are already thin.

At present, Bangladesh dementia numbers keep climbing - about 1.1 million people live with it in 2020. Experts say those numbers will rise, reaching 1.37 million soon after 2025, and by 2041, that count may reach 2.41 million. However, there is still a noticeable lack of technological support for early detection and real-time communication assistance. Most care paths center on diagnosis and extended medical stays. Useful voice-driven aids for spotting mental changes or navigating everyday chats aren't around enough. The absence hits harder where resources are tight. In poorer regions, seeing experts or using complex tests becomes even more difficult. Without simple tools, families face delays when help should arrive fast.

Aiming to spot dementia earlier, this study develops a smart tool for analyzing voice data. During conversation, if someone stumbles over words or finishes speaking mid-sentence, help arrives fast through predicted phrases. Instead of waiting, the system jumps in - guided by pairings of voice tech and grammar forecasting methods. Smooth talk flows more easily when glitches are caught early, which also lightens the burden on caregivers.

Key research questions guiding our investigation include:

- Can dementia be effectively detected through voice-based analysis?
- How can speech recognition and sentence prediction models assist dementia patients during real-time conversations?
- To what extent can such a system reduce communication barriers and frustration among patients and caregivers?

This research has the potential to contribute to technological and social sectors. If we look at it from a technological perspective, it introduces a speech-based system designed to detect early signs of dementia and assist patients in real time, whereas, socially, it aims to improve the quality of everyday communication, ease the emotional and practical burden on caregivers, and encourage timely intervention. It will tackle communication challenges directly. Furthermore, the system could help create a more supportive and understanding environment for patients, especially in countries that have limited resources, such as Bangladesh.

Dementia is a growing public health concern in Bangladesh and worldwide, making early detection and communication support increasingly important. This study proposes a voice-based system using speech analysis and predictive technologies to aid patients, support caregivers, and improve dementia care in low-resource settings.

## 2- Literature Review

Most existing studies on detecting dementia tend to focus primarily on widely spoken languages such as English. As a result, languages like Bangla are often overlooked and receive significantly less attention in this area. A crucial limitation in current research is the absence of studies that utilize Bangla voice data for dementia identification. The majority of previous studies rely on datasets such as Pitt Corpus (DementiaBank), which contain English speech recordings [10][11]. It is a publicly available dataset containing speech recordings and corresponding transcripts from Alzheimer's disease patients and healthy elderly individuals, primarily collected through picture description tasks to analyze cognitive impairments associated with dementia [2]. For instance, Kumar et al. (2022) explored Machine Learning (ML) and Deep Learning (DL) approaches on the DementiaBank dataset [11]. However, a model trained on English speech may not accurately capture Bangla-specific prosodic patterns such as variations in intonation, syllable timing and word articulation. As there is no publicly available dataset of recorded Bangla speech, this study takes an innovative approach by collecting a dedicated dataset from Bangla-speaking participants. Additionally, we transcribe the collected voices into text to facilitate better analysis and improve dementia detection using linguistically-based features. The transcription of Bangla voices remains particularly challenging compared to widely spoken languages like English, making it a crucial aspect of this research. In addition, no prior studies have attempted to create sentence auto-completion systems for dementia patients, specifically in Bangla, to address their communication challenges. While research on Bangla next-word prediction and sentence completion exists, such as the study that proposed a Bi-LSTM model trained on a 1.7 GB dataset from various Bangla news sites. Notably, their research focuses on general language modeling, whereas this research is specially designed to assist dementia patients [9].

The application of deep learning methods has opened new ways for identifying dementia through voice recordings. A proposed deep learning model using voice recordings from the Framingham Heart Study, consisting of 1,264 audio samples from 656 participants across cognitive states [28]. They extracted MFCCs from raw audio and fed them into a hierarchical LSTM and 1D CNN. CNN achieved the best performance with an AUC of 0.805. The study "Computer-Based Evaluation of Alzheimer's Disease and Mild Cognitive Impairment" by Hernández-Domínguez in 2018, utilizes the most common Pitt Corpus from the DementiaBank dataset for dementia analysis [7]. The research focuses on assessing Alzheimer's disease (AD) and mild cognitive impairment (MCI) patients, analyzing 517 transcriptions (257 AD, 217 healthy controls, and 43 MCI). The methodology involves extracting linguistic and phonetic features, including vocabulary richness, part-of-speech usage and MFCCs. These features are followed by classification using Support Vector Machines (SVM) and Random Forest (RF) models. This study achieves 81% sensitivity with an AUC of 0.79 for distinguishing AD from healthy controls, improving to 85% sensitivity with an AUC of 0.76 when including MCI patients. Similarly, the study mentioned earlier by Kumar used the same Pitt corpus dataset, comprising 442 audio recordings [11]. In this study, ML and DL Models were applied to classify dementia using key acoustic and speech-based features. The extracted features were MFCCs, Gammatone Cepstral Coefficients (GTCCs), jitter, shimmer and formants. Among the ML models, Random Forest achieved the highest accuracy of 87.6%. On the other hand, the PRCNN model outperformed other DL approaches, achieving 85% accuracy, demonstrating the potential of Deep Learning for dementia detection. Additionally, another study on early dementia detection leveraged the same dataset, but it analyzed 1,541 text samples and extracted linguistic and semantic features, including hesitation counts, type-token ratio (TTR), and embeddings (Word2Vec, GloVe, Sentence embeddings) [10]. Various ML models, including RF, K-Nearest Neighbors (KNN), SVM and Artificial Neural Networks (ANN). The RF results achieved 99.35% accuracy using Word2Vec embeddings, while the ANN results reached 99.67% accuracy with the same feature set. Beyond speech-only approaches, multimodal analysis has gained significant attention in dementia detection. Since our study also adopts a multimodal approach, it aligns with this research trend. One notable study by Campbell [3], "Alzheimer's Dementia Detection from Audio and Text Modalities," investigated both speech and text-based features. This research used the ADReSS Challenge dataset, a subset of the DementiaBank Pitt Corpus [13]. The dataset contains 108 speech recordings and transcripts (54 AD, 54 non-AD). The research explores four speech-based models (i-vectors, x-vectors, statistical functionals, and speech fluency analysis) and two text-based models (an RNN with GloVe and linguistic-feature-based SVMs). Text-based models outperform speech-based models, with RNN achieving 79.6% accuracy. Another study applies NLP to detect cognitive impairment from clinical notes. Using a dataset of 16,428 patients with 8,656 annotated sequences, the study compares TF-IDF and Clinical BERT models [23]. The TF-IDF model achieved 84% accuracy, while Clinical BERT significantly outperformed it with 93% accuracy. No direct research has yet focused on detecting dementia using Bangla speech or text. However, some studies on speech emotion recognition offer valuable insights. Two such works by Shruti et al. [18] in 2023 and Sultana et al. [19] in 2022 both use the SUBESCO dataset (7,000 samples labeled with seven emotions). The first study combined Mel-spectrograms with features such as ZCR, Chroma, and RMS and achieved 90% weighted accuracy using RF. The second introduces a deep learning-based model (CNN+TDF+BLSTM) that achieves 86.86% accuracy. Lastly, there is a lack of visible research on sentence completion in Bangla, especially for dementia patients. To address this, Sarker et al. propose a hybrid model for word completion and an LSTM with N-gram ranking for sentence prediction in 2020 [17]. Two custom datasets were 1.24M words and 3M words from the Bangla news portal. Features included one-hot-encoded word sequences, and the results showed 84% accuracy on the first dataset and 81% on the second. Another significant research fine-tunes WangchanBERTa, a Thai-specific RoBERTa model, to classify this sentence using a labeled dataset of 2,000 sentences [15]. The model achieves 99.65% accuracy with a lower computational cost. Still, none of these studies are directly aligned with our approach, which focuses on sentence completion in a different linguistic and contextual setting.

This study addresses several critical gaps in dementia detection and communication assistance.

- Multimodal dementia detection by combining acoustic features from voice data with linguistic patterns from transcribed text. Additionally, it introduces a sentence autocompletion model to assist dementia patients in real-time communication.
- Addressing the scarcity of dementia research in Bangla, this study designs and collects a custom dataset specially tailored to the cultural context of Bangladesh.
- A transcription task is performed to convert voice recordings into text, ensuring consistency across modalities and enabling systematic analysis.
- Despite the challenges posed by a limited dataset, this study demonstrates the feasibility of using LSTM and Bi-LSTM models for dementia related tasks and achieved competitive performance.

## 3- Methodology

Table 1 presents a summary of the collected dataset, including 740 voice samples (370 from dementia patients and 370 from non-dementia individuals) and 1,890 sentences, comprising 740 transcribed sentences and 1,150 additional sentences used for the sentence completion task. Demographic details, including age and gender distribution, are also provided.

**Table 1:** Summary of Dataset Collected from Interviewed Participants.

| Category | Dementia | Non-Dementia | Total |
|---|---|---|---|
| Voice Samples | 370 | 370 | 740 Voices |
| Age-Range | (62-78) years | (22-35) years | |
| Text Sentences | 1520 (Transcribed + Additional) | 370 (Transcribed) | 1890 Sentences |

### *3-1- Dataset Collection*

This research was designed to utilize a dataset comprising voice recordings to analyze and detect dementia while providing sentence autocompletion support. The dataset collection process was focused on ensuring diversity, relevance and alignment with daily communication patterns. The sentences were designed around the five basic human needs: food, clothing, shelter, education and healthcare, reflecting common communication scenarios encountered in daily life.

Voice data were collected from 117 individuals with dementia and 155 without dementia, including both male and female participants. For the dementia group, participants were selected based on their age, clinical condition and cognitive assessment using the Mini-Mental State Examination (MMSE), with scores of 23 or below indicating cognitive impairment, consistent with established clinical thresholds [29]. Standard exclusion criteria, such as severe psychiatric disorders, delirium, language barriers and low educational background that could significantly affect cognitive assessment were considered during participant selection. The dementia group consisted of patients with clinically diagnosed dementia under continuous medical supervision who exhibited characteristic speech patterns, including frequent pauses, hesitation, and difficulty recalling words. In contrast, the non-dementia participants showed no observable dementia-related symptoms in their speech. The non-dementia participants were selected based on normal daily functional activities and MMSE scores of 24 or above, indicating normal cognitive function.

The age ranges for the dementia and non-dementia groups were 62 to 78 years and 22 to 35 years, respectively, reflecting the age range of the participants in this study. Dementia is predominantly associated with older adults, affecting approximately 5% of individuals aged 65-74 and more than 30% of those aged 85 and above [30]. Therefore, most participants in this study with dementia belong to the older age group. Each participant was asked to read the prepared sentences to ensure consistent linguistic content and controlled recording conditions. Although reading tasks may not fully reflect spontaneous conversational speech, the sentences were designed to reflect typical daily communication patterns and to allow reliable comparisons of acoustic and linguistic features between groups.

A total of 740 voice recordings were collected, with 370 from dementia patients and 370 from non-dementia individuals. All recordings were saved in WAV format with a sampling rate of 44 kHz to ensure high-quality audio. The voice recordings were transcribed into text for linguistic analysis. To enhance the sentence autocompletion task, an additional 1,150 dementia-related sentences were curated and combined with the 370 transcribed dementia sentences, resulting in a total of 1,520 dementia-related sentences and an overall dataset of 1,890 sentences. Ethical considerations, including informed consent and data confidentiality, were strictly adhered to during data collection.

### *3-2- Recorded File Analysis Strategy for Interviewed Participants*

The process of detecting dementia and completing sentences in Bengali is designed to be structured and systematic. The workflow, as illustrated in Figure 1, comprises several stages, each playing a critical role in achieving the research objectives. The process begins with data collection, where audio recordings of participants are generated. These recordings are then subjected to audio preprocessing to enhance the consistency of the audio signals. Subsequently, feature extraction is performed on the preprocessed audio to identify acoustic features associated with dementia or non-dementia. Subsequently, the audio recordings are transcribed into text using a Bangla automated speech recognition (ASR) system. The transcribed sentences are then matched with the original sentences to improve transcription accuracy. The transcribed text undergoes text preprocessing, followed by feature extraction to identify linguistic patterns and semantic features. These features are then used to implement classification models that distinguish between sentences.

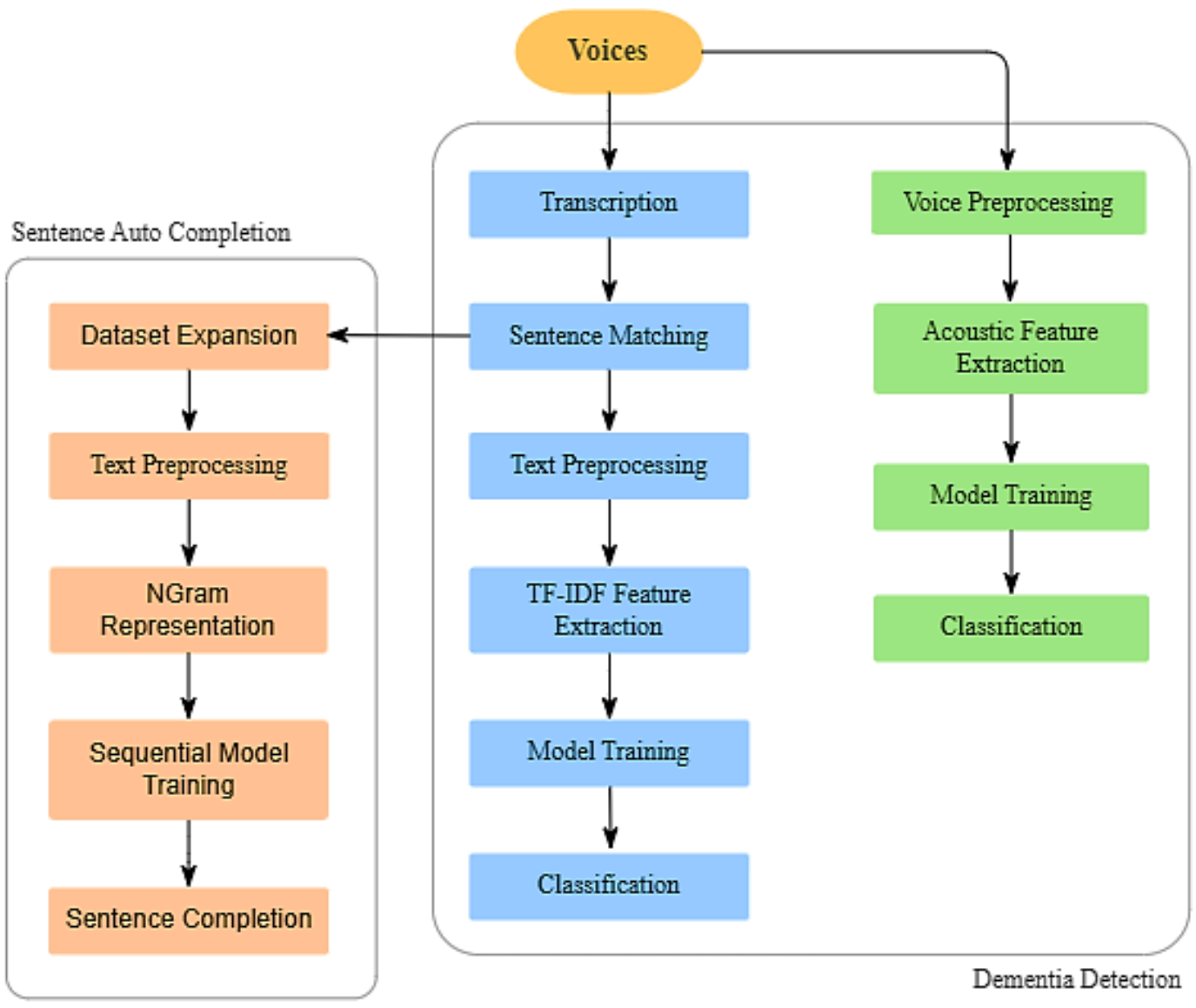

**Figure 1:** Systematic process for the proposed task.

For the sentence autocompletion task, the transcribed dementia sentences are augmented with additional sentences to create a robust dataset. N-gram sequences are extracted from dementia sentences to capture contextual patterns. These sequences are used to train models that predict the next word in a sentence, enabling sentence completion. Overall, this methodology provides a comprehensive, reproducible framework for detecting dementia through voice analysis and for performing sentence autocompletion. The structured workflow, as depicted in Figure 1, ensures that each step builds upon the previous one, maintaining consistency and reliability throughout the research process.

### *3-3- Acoustic Analysis*

#### *3-3-1. Voice preprocessing and feature extraction*

Cognitive and linguistic impairments associated with dementia frequently manifest as noticeable modifications in an individual's speech patterns. Increased pauses, reduced fluency, and changes in melody and intonation are frequently the result of these challenges. For this study, we compiled 740 raw voice samples from interviews (370 dementia and 370 non-dementia) and systematically processed them to enhance signal quality. The soundfile library was utilized to process speech data in WAV format throughout the pipeline. Samples were recorded at a sampling rate of 44.1 kHz to ensure comprehensive capture of the full acoustic spectrum, including high-frequency harmonics and articulatory details. Spectral subtraction and band-pass filtering (20 Hz-8 kHz) were employed to reduce background noise. Subsequently, the samples were downsampled to 22.5 kHz to reduce computational demands while retaining diagnostic relevant acoustic information. Spectral noise gating consists of first estimating a noise profile from the parts of the speech where non-speech segments were dominant. This forms the audio parts with low energy and then removes non-speech segments according to the previously estimated noise profile. As illustrated in Figure 2, this effectively reduced background noise in both dementia and non-dementia speech waveforms while retaining diagnostically critical features. Following spectral subtraction, a technique that identifies and removes the noisiest parts of a sound using statistical analysis. Conclusively, adaptive smoothing was applied to the filtered signal to preserve temporal continuity and the integrity of vocal patterns. Amplitude normalization was implemented to eliminate volumetric disparities across the samples, ensuring consistent baseline loudness. These processing stages collectively yielded a homogenized, acoustically sanitized dataset, optimized to facilitate robust feature extraction.

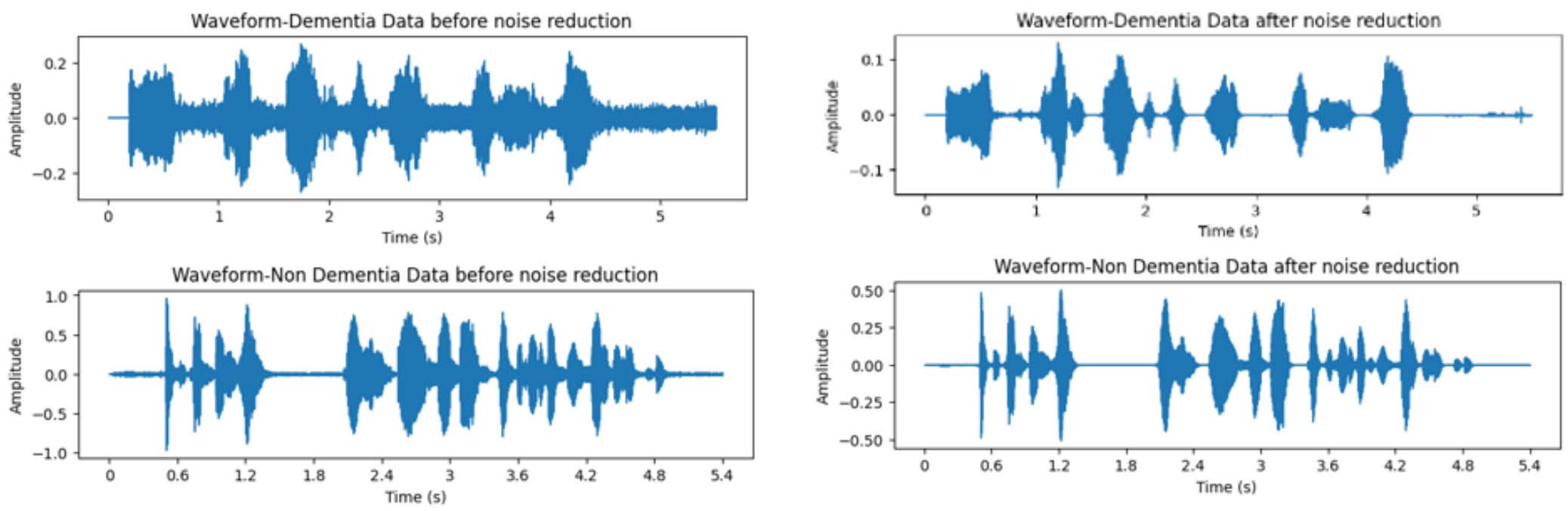

**Figure 2:** Waveform visualization before and after noise reduction.

The feature extraction phase is a crucial step in analyzing speech patterns for dementia identification. Because dementia can significantly affect speech fluency and articulation, this process involves identifying and extracting key acoustic features from recorded speech samples. These features help in understanding the changes and irregularities in speech that may highlight the presence of dementia. One of them is MFCC (Mel-Frequency Cepstral Coefficients). MFCC represents the short-term power spectrum of sound by converting frequencies to the Mel scale, which mimics human auditory perception. A total of 13 MFCC coefficients were extracted to model spectral properties. MFCCs represent the short-term power spectrum of sound by mapping frequencies to the Mel scale, which approximates human auditory perception. This property makes MFCCs effective for capturing significant spectral attributes of speech. Articulation of a phoneme is a process that involves synchronous motion of the tongue, lips, and jaw. In individuals with cognitive decline, disruptions in neural mechanisms responsible for motor planning and speech control can affect articulatory precision and phoneme production. As MFCCs model the spectral envelope of speech, they can capture variations in pronunciation, articulation, and vocal quality.

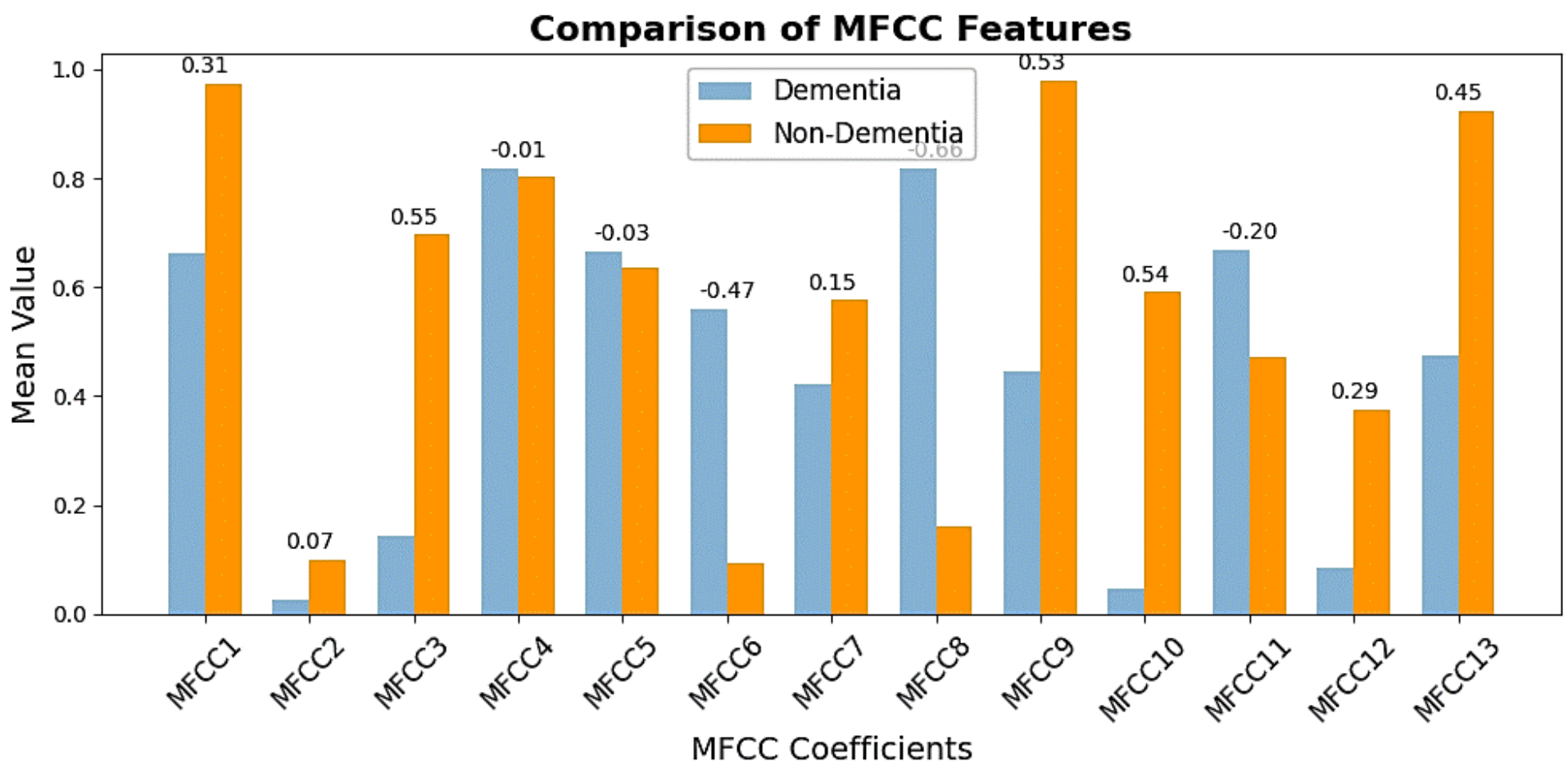

**Figure 3:** Significant MFCC coefficients for both voice types.

To visualize the relationship between dementia and non-dementia frequencies, Figure 3 illustrates a comparative analysis of MFCCs averaged across the dementia and non-dementia cohorts, revealing noticeable differences in several coefficients, specifically MFCC1, MFCC3, MFCC6, MFCC8, MFCC9, MFCC12, and MFCC13. To differentiate between these two groups, these variations suggest that specific coefficients could play a key role in identifying speech patterns associated with dementia, offering potential insights. Other features, including spectral bandwidth, spectral centroid, and Zero-Crossing Rate (ZCR), were examined to quantify the irregularities of articulation.

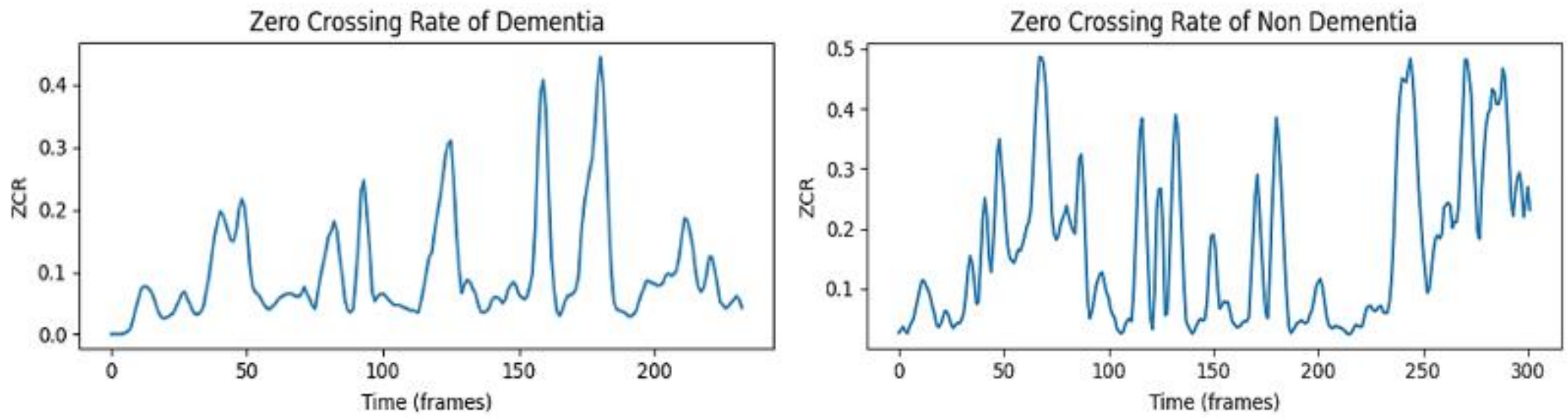

**Figure 4:** Comparison of ZCR in dementia and non-dementia samples.

The ZCR (Zero Crossing Rate) showed unique acoustic characteristics of dementia and non-dementia speech samples. According to Figure 4, speech in the non-dementia class showed a higher peak ZCR magnitude (nearly 0.5), indicating greater overall vocal variability and dynamic prosody, including rapid pitch variations and high articulatory precision. These observations highlight ZCR as a measurable indicator of compromised temporal and spectral properties. The observations above indicate that ZCR is a quantifiable marker that signifies impaired temporal and spectral activity in dementia speech. The dementia samples of lower ZCR suggest loss of temporal variability and voicing stability. The ZCR measures the rate at which the speech signal changes sign within a given time frame and is calculated as:

$$zcr = \left(\frac{1}{T}\right)\sum_{t=0}^{T-1}[\![s_t s_{t-1} < 0]\!] \tag{1}$$

Where $T$ represents the total number of samples within the analyzed frame of the signal. In this expression, $s_t$ denotes the signal amplitude at the current sample $t$, while $s_{t-1}$ represents the signal amplitude at the previous sample time $t-1$. The term $s_t s_{t-1} < 0$ is an indicator function that evaluates whether a zero-crossing has occurred between two consecutive samples. It takes the value 1 when the product. $s_t s_{t-1}$ is less than zero, indicating that the signal has changed sign, and 0 otherwise. Prosodic features, such as Chroma shifts, can explain speech rhythm, pitch, and fluency, which are often affected in dementia-related voices. Chroma features map frequencies to 12 chroma-based pitch representations, capturing harmonic content rather than absolute pitch. Jitter and Root Mean Square (RMS) energy were incorporated as acoustic features because they capture complementary aspects of vocal impairments associated with cognitive decline. Jitter measures cycle-to-cycle variations in the fundamental frequency and reflects the stability of vocal fold variation. In dementia, it often leads to increased pitch instability, making jitter a sensitive indicator of phonatory irregularities. In contrast, RMS energy represents the short-term intensity of speech and is related to vocal effort and respiratory support. Cognitive impairments may affect breath control and speech production, resulting in reduced or inconsistent energy patterns. Together, jitter (frequency instability) and RMS energy (amplitude variation) provide complementary information about control and vocal intensity, enhancing the robustness of speech-based classification models.

#### *3-3-2. Speech classification model implementation*

A hybrid framework integrating traditional machine learning and deep learning models was developed to distinguish between dementia and non-dementia speech patterns. Logistic Regression was utilized with L2 regularization to avoid overfitting. A Random Forest model combining 100 decision trees was implemented to improve classification accuracy. Additionally, Gradient Boosting was employed to iteratively minimize errors and decrease bias by sequentially refining weak learners, improving overall performance. In this research, a Long Short-Term Memory (LSTM) network has been implemented to classify dementia based on speech analytics. This model architecture used a single LSTM layer with 128 units to capture temporal dependencies in the speech dataset, including pitch variables and pauses, which are important for distinguishing dementia-related patterns. A dropout layer (rate=0.2) was implemented, randomly deactivating neurons during training to prevent overfitting. The output layer employed a dense layer with a sigmoid activation function for binary classification. Features were standardized and reshaped into 3D tensors to align with LSTM input requirements. Labels were one-hot encoded to ensure compatibility**.** The Adam optimizer was used to optimize the model, with categorical cross-entropy as the loss function and accuracy as the evaluation metric. Cross-validation was used to assess generalizability, and early stopping was used to halt training if validation loss remained constant. This was trained with a batch size of 64, and early stopping was used based on the validation loss to avoid overfitting. For all implemented models, the dataset was split into training and test sets at 80:20, with 80% allocated to training and 20% to evaluation. The effectiveness of all models was evaluated using standard metrics, including classification accuracy, precision, recall, and F1 score. In total, these metrics validated the robustness of the framework in classifying dementia and non-dementia individuals.

### *3-4- Whisper-based speech transcription*

The transcription pipeline utilizes the Whisper model, a transformer-based automatic speech recognition (ASR) system from the Hugging Face ecosystem, fine-tuned for Bangla transcription [16]. The model operates on GPU-accelerated hardware for optimal performance and consists of a feature extractor that converts raw audio into log-Mel spectrograms, a tokenizer, a processor configured for Bangla, and a sequence-to-sequence transformer for text generation [12]. Audio files in WAV format were pre-processed by resampling to 16 kHz and converting into numerical representations compatible with the model's input requirements. The transcription process involves transforming the audio signal. $x(t)$ into a log-Mel spectrogram $L(m, l)$, which is processed by a transformer-based encoder-decoder architecture. The encoder generates hidden states. $H$, and the decoder produces token sequences $y$ autoregressively. The final transcription is obtained by decoding y into text ($\hat{T}$), with special tokens removed for clean output.

$$L(m, l) = MelSpec\left(x(t)\right) \tag{2}$$

$$H = Encoder\left(L(m, l)\right) \tag{3}$$

$$y = argmax\ P(y|H) \tag{4}$$

$$\hat{T} = Decoder\ (y) \tag{5}$$

After successfully transcribing the voices into text, some words were inaccurate, which could affect the model's performance, as evidenced in Table 2. The table presents a comparison between the original recorded sentences and their transcriptions, along with the corresponding labels. In the first sentence, the first-person pronoun "আমি" (I) was incorrectly transcribed as the third person "তিনি" (he/she). Similarly, in the second sentence, "তোমার মা" (your mother) was transcribed as "মার মা" (mothers' mom). While these errors alter the original meaning, the transcribed version remains semantically coherent. However, more problematic cases emerged when the sentences, like the third sentence, where words were transcribed into non-existent or meaningless forms (e.g., আড়ামের, বৈপড়া). To address these inaccuracies, matched the transcribed sentences with the main designed sentences from which the voices were originally recorded. Using a custom correction function, we compared each transcribed word with its corresponding word in the correct sentence and replaced any mismatches based on a similarity threshold of 0.8. This ensured that the final transcriptions were as accurate as possible. After that, the audio files were organized into Dementia and Non-Dementia categories, and the transcriptions, along with metadata (label and class), were stored in a CSV file for downstream analysis, enabling seamless integration with subsequent tasks.

**Table 2:** Audio sample transcription results.

| Recorded sentences | Transcribed output | Class |
|---|---|---|
| আমি কি খেয়েছি মনে করতে পারছি না।<br>*(I can't remember what I ate)* | তিনি কী খেয়াসি মনে করতে পাচ্ছি না।<br>*(He/She can't remember what he/she ate)* | Dementia |
| তোমার মা এখন মাঝেমধ্যে আমাকে খাইয়ে দেয়।<br>*(Your mother now sometimes feeds me)* | মার মা এখন মাঝে মধ্যে আমাকে খাইয়ে দায়।<br>*(Mother's mom now sometimes feeds me)* | Dementia |
| কাজ শেষে আরামের জন্য বই পড়া আমার পছন্দের।<br>*(After work, I like reading books for relaxation)* | কাজ শেষে আড়ামের জন্য বৈপড়া আমার পছন্দের।<br>*(After work, I like 'vipara' for 'aram')* | Non-Dementia |

### *3-5- Linguistic Analysis*

#### *3-5-1. Text preprocessing and feature extraction*

The text processing pipeline begins with merging transcribed sentences from both the dementia and non-dementia categories into a single dataset. This unified dataset undergoes initial cleaning to remove unnecessary elements, such as Bangla periods (।) and commas (,), which do not contribute to the semantic meaning of the sentences. Specifically, sentences are truncated at the first occurrence of a Bangla period, and all commas are eliminated to ensure consistency in the text structure. Next, a predetermined list of Bangla stop words is applied to remove common function words that do not add any meaningful content. Additionally, single-character fragments and non-Bangla patterns are filtered out to retain only linguistically valid tokens. Any empty strings resulting from the cleaning process are replaced with a placeholder (EMPTY) to maintain dataset integrity. Finally, the processed dataset containing clean, normalized text is saved for feature extraction and classification. The feature extraction process involves transforming the cleaned text data into numerical features using the Term Frequency-Inverse Document Frequency (TF-IDF) metric. This statistical measure evaluates the importance of words in a document relative to a corpus. A custom tokenizer was implemented to split Bangla text into tokens based on spaces, ensuring proper handling of the language's structure. The TF-IDF vectorizer was configured to capture unigrams, bigrams and trigrams (1–3-word sequences), enabling the model to account for both individual words and contextual phrases. To maintain computational efficiency and focus on the most discriminative features, the vectorizer was limited to the top 3,000 features ranked by TF-IDF score. The cleaned text data was then transformed into a TF-IDF matrix, where each row corresponds to a document (sentence) and each column corresponds to a feature (word or n-gram).

#### *3-5-2. Text classification model implementation*

To detect Dementia from the extracted features, three machine learning and one deep learning models were implemented. Each model was designed, trained and evaluated to classify text data into both categories.

Logistic Regression, a linear classification model, was employed due to its simplicity and effectiveness in binary classification tasks. The model was initialized with L2 regularization (C=0.1) to prevent overfitting and configured with class balancing. The dataset was split into training (80%) and testing (20%) sets, ensuring stratified sampling to maintain class distribution. An ensemble-based Random Forest model was implemented to leverage its ability to handle non-linear relationships and reduce overfitting. The model was configured with 100 decision trees and no limit on tree depth to capture complex patterns in the data. The dataset was split similarly to the previous model. The Support Vector Machine is a robust classifier with a linear kernel that effectively handles high-dimensional TF-IDF features. The target variable was encoded numerically, and the dataset was split into the same ratio as other models. The model was trained to find the optimal hyperplane that maximally separates the two classes, leveraging regularization to prevent overfitting. The models were trained on TF-IDF features, and a fixed random seed (random state = 42) was used for reproducibility.

**Table 3:** LSTM model summary.

| *Layer (type)* | *Output Shape* | *Param #* |
|---|---|---|
| Embedding | (None, max_len, 60) | Vocab_size * 60 |
| LSTM | (None, 128) | 96,768 |
| Batch Normalization | (None, 128) | 512 |
| Dense (ReLU activation) | (None, 64) | 8,256 |
| Dropout | (None, 64) | 0 |
| Dense (Sigmoid activation) | (None, 1) | 65 |
| | Total Parameter | 105,857 |

The Long Short-Term Memory (LSTM), a deep learning model, was implemented for dementia detection using text data. The model leverages sequential data processing to capture temporal dependencies in the text. The text had been preprocessed by tokenizing sentences into sequences of integers and padding them to a fixed maximum sequence length (L=13) to ensure uniform input dimensions. The model architecture specifications detailed in Table 3 included an embedding layer to convert tokens into a dense vector of dimension d = 60, an LSTM layer with 128 units, dropout (rate=0.2), L2 regularization to prevent overfitting, and batch normalization for training stability. A fully connected layer with 64 units and ReLU activation was added, along with a dropout layer (rate=0.5) to reduce overfitting further.

The output layer used a sigmoid activation for binary classification. As shown in the table, the LSTM architecture employs three gating mechanisms: The forget gate ($f_t$), input gate ($i_t$) and output gate ($o_t$) to control the flow of information. While the cell state ($C_t$)) and hidden state ($h_t$)) capture long-term dependencies. The embedding layer maps tokenized sequences into a dense vector ($E = W_e.X$). The model was compiled with the Adam optimizer and trained using binary cross-entropy loss [8], defined as:

$$\mathcal{L} = -\frac{1}{N}\sum_{i=1}^{N}[(y_i \log(\hat{y}_i) + (1 - y_i)\log(1 - \hat{y}_i))] \quad (6)$$

where $y_i$ is the true label, $\hat{y}_i$ is the predicted label, and $N$ is the number of samples. Training was conducted for 20 epochs with early stopping to restore the best weights. Non-linear activation functions, such as the sigmoid function $\left(\sigma(z) = \frac{1}{1+e^{-z}}\right)$ is used in the output layer for binary classification, while the ReLu function ($ReLU(z) = \max(0, z)$) is applied in the fully connected layer to introduce non-linearity. These activation functions ensure the model can effectively capture complex patterns in the text data.

### *3-6- Sentence Completion Model*

The sentence autocompletion task aims to assist dementia patients by predicting the next word in a sequence. To enhance the model's ability to assist dementia patients effectively, the dataset was expanded beyond the transcribed sentences used for classification. This expansion comprises Bangla sentences that reflect the daily experiences and common conversations of elderly individuals, particularly those affected by dementia. These sentences were carefully crafted to cover a wide range of topics related to basic human needs, ensuring comprehensive coverage of everyday communication scenarios. The sentences were preprocessed and prepared for training a Bidirectional-LSTM model. The preprocessing pipeline began with cleaning the text by removing punctuation marks such as Bangla periods ( ।), commas (,), exclamatory marks (!), and question marks (?) to ensure uniformity. Each sentence was appended with a special <end> token to mark the end of the sequence, enabling the model to learn sentence boundaries effectively. A tokenizer was initialized with an out-of-vocabulary <oov> token to handle unseen words during inference. The sentences were tokenized into sequences of integers, creating a vocabulary of unique words. The vocabulary size (V) was calculated as $V = len(tokenizer.word_index) + 1$, where the additional token accounts for padding. To generate training data, n-gram sequences were created from the tokenized sentences. For a sentence $S = [w_1, w_2, \dots, w_n]$, sequences of increasing length generated as $[w_1, w_2]$, $[w_1, w_2, w_3]$, $[w_1, w_2, w_3, w_4]$, and so forth. Each sequence follows a predictive framework where all preceding tokens serve as contextual inputs, while the final token ($w_i$) acts as the target label. This structure is fully specified in Table 4, which shows the sentence-to-sequence generation and the explicit mapping of predictor tokens and label tokens. The sequences were padded to a fixed length (L) using pre-padded to ensure all inputs have the same dimensions, which is necessary for training the model.

**Table 4:** N-gram sequence overview.

| *Predictors (x)* | *Predictors token (x)* | *Labels (y)* | *Labels token(y)* |
|---|---|---|---|
| আমি খাবার<br>*(I foods)* | 6 52 | কোথাও<br>*(anywhere)* | 185 |
| আমি খাবার কোথাও<br>*(I foods anywhere)* | 6 52 185 | খুঁজে<br>*(searching)* | 112 |
| আমি খাবার কোথাও খুঁজে<br>*(I foods anywhere searching)* | 6 52 185 112 | পাচ্ছি<br>*(finding)* | 133 |
| আমি খাবার কোথাও খুঁজে পাচ্ছি<br>*(I foods anywhere searching finding)* | 6 52 185 112 133 | না<br>*(not)* | 3 |
| *আমি খাবার কোথাও খুঁজে পাচ্ছি না*<br>*(I foods anywhere searching finding not)* | 6 52 185 112 133 3 | <end> | 2 |

After padding, the sequences were split into predictors (x) and labels (y). Then they were converted into one-hot encoded vectors. In this representation, each label $y_i$ is transformed into a vector of size V, where all elements are 0 except for a 1 at the index corresponding to the target word. For example, if the target word is the 5th word in the vocabulary, the one-hot encoded vector would be $y_i = [0, 0, 0, 0, 1, 0, \dots, 0]$. This one-hot encoding enables the model to perform multi-class classification and predict the next word from the entire vocabulary.

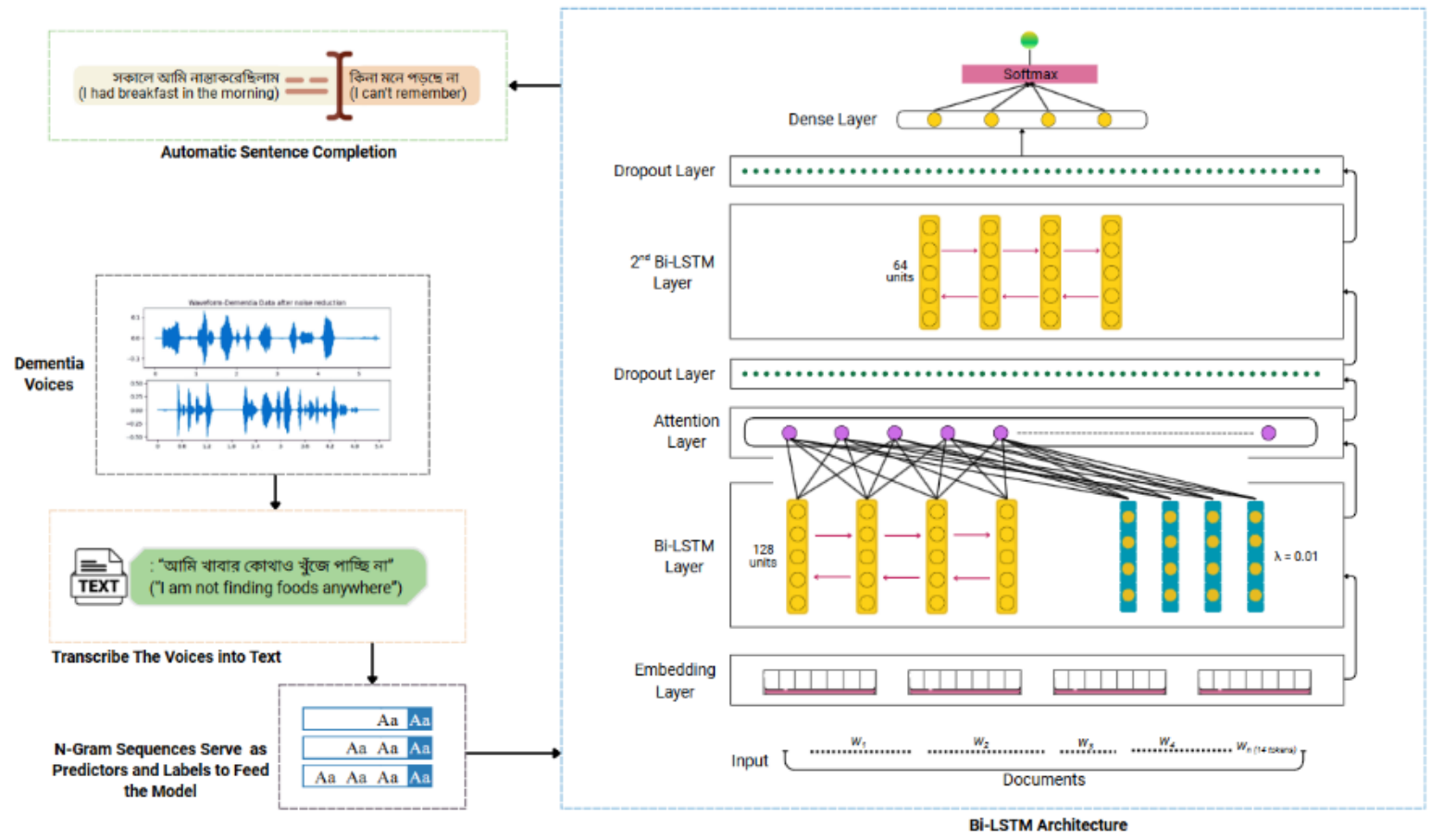

**Figure 5:** Sentence completion process with a Bi-LSTM model architecture.

The Bidirectional LSTM (Bi-LSTM) architecture is enhanced with an attention mechanism, which captures contextual information from both past and future words in a sentence. The preprocessed text data, which consists of Bangla old-age-specific sentences, was split into predictors and labels. The dataset was divided into training (60%), validation (20%), and testing (20%) sets to ensure robust evaluation and prevent overfitting. The model architecture in Figure 5 begins with an input layer that accepts sequences of fixed length. These sequences are passed through an embedding layer, which maps each token to a dense vector of size 128, capturing semantic relationships between words. The embedded sequences are then processed by a Bidirectional LSTM layer (128 units) with L2 regularization (\lambda=0.01), which analyzes the text in both forward and backward directions to capture contextual information. An attention mechanism is applied to focus on the most relevant parts of the sequence, improving the model's ability to handle long sentences. A dropout layer (rate=0.3) is applied to prevent overfitting, followed by a second Bi-LSTM layer (64 units), which further refines the sequence representations. Another dropout layer (rate=0.3) is added before the final dense layer. The output layer uses a SoftMax activation function to predict the next word from the vocabulary. The model was compiled using Adam optimizer (learning rate=0.0005) and sparse categorical cross-entropy loss, which is suitable for multi-class classification. It was trained for 60 epochs on the training set, with the validation set used to monitor performance and ensure generalization.

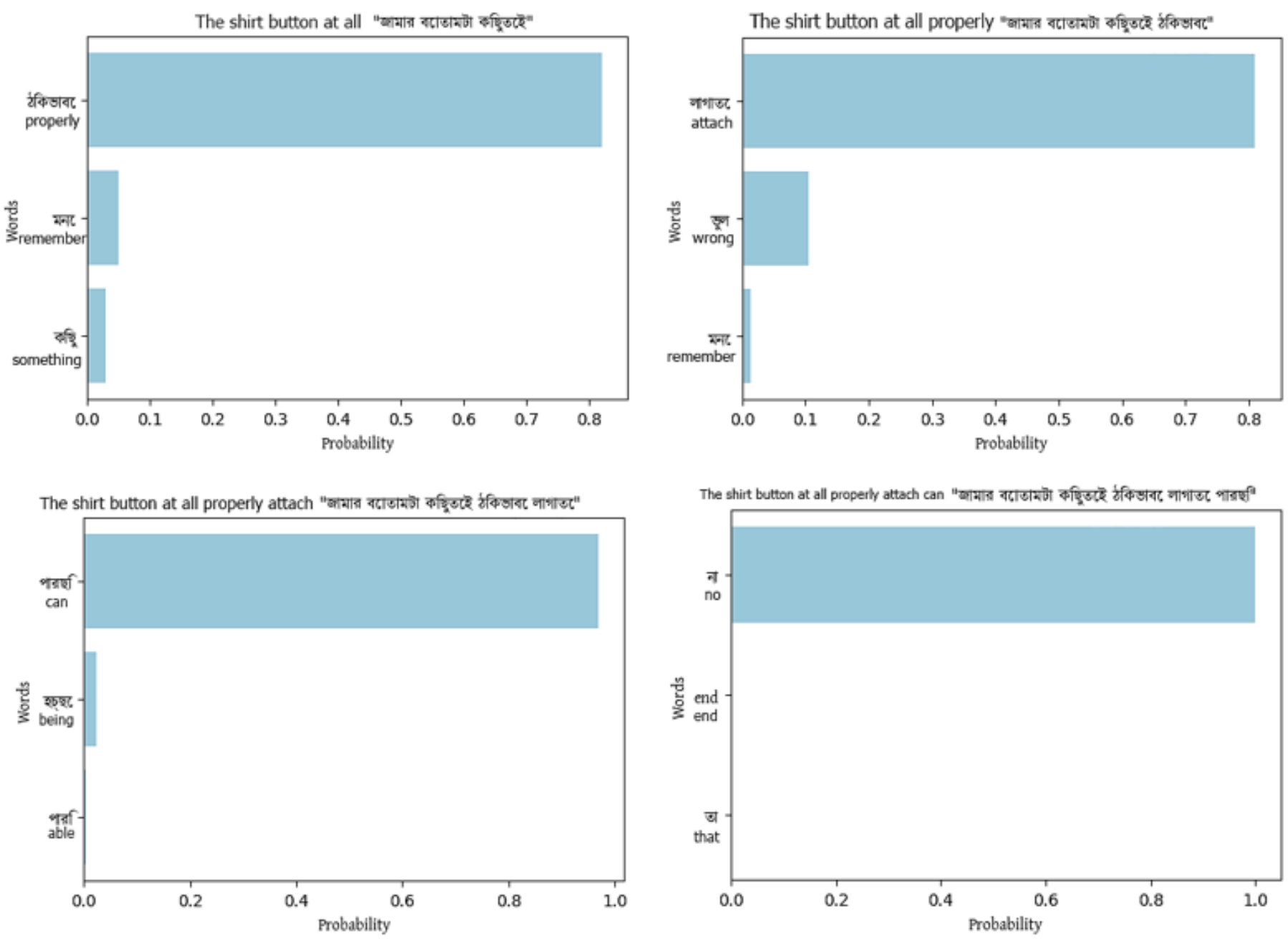

**Figure 6:** Sequential prediction of the next word based on probability.

The model predicts the entire sentence by iteratively generating the next word based on the seed text. The process begins with an initial seed text, which is tokenized and padded to match the input sequence length (L-1). The model predicts the next word by selecting the word with the highest probability from the SoftMax output, as illustrated in Figure 6. Given the seed text "জামার বোতামটা কিছুতেই" *(translated as "The shirt button at all")*, the model generates potential next words based on probability scores. Among the suggested words, "ঠিকভাবে" *(properly)*, "মনে" *(remember)*, "কিছু" *(something)*, the word "ঠিকভাবে" *(exactly)* achieved the highest probability of 82%. Following this, the model predicts the next word, “লাগাতে” (attach), with 80% probability. Similarly, the subsequent words “পারছি” *(can)* and “না” *(not)* are both assigned a probability score of 97%, leading to the completion of the sentence as “জামার বোতামটা কিছুতেই ঠিকভাবে লাগাতে পারছি না” *(Natural English translation: I cannot attach the shirt button properly at all)*. The process repeats until the model predicts the special <end> token, signalling the end of the sentence. The model was evaluated using training, validation, and test accuracy metrics to assess predictive performance. Additionally, perplexity was computed to measure the model’s confidence and uncertainty in next-word prediction, further demonstrating its effectiveness in generating meaningful, contextually appropriate sentence completions.

## 4- Results and Analysis

### *4-1- Performance Evaluation of Voice-Based Classification*

In this study, we explored several machine learning methods to distinguish between dementia and non-dementia from speech data, and the results underscored the remarkable performance of LSTM (Long Short-Term Memory) networks.

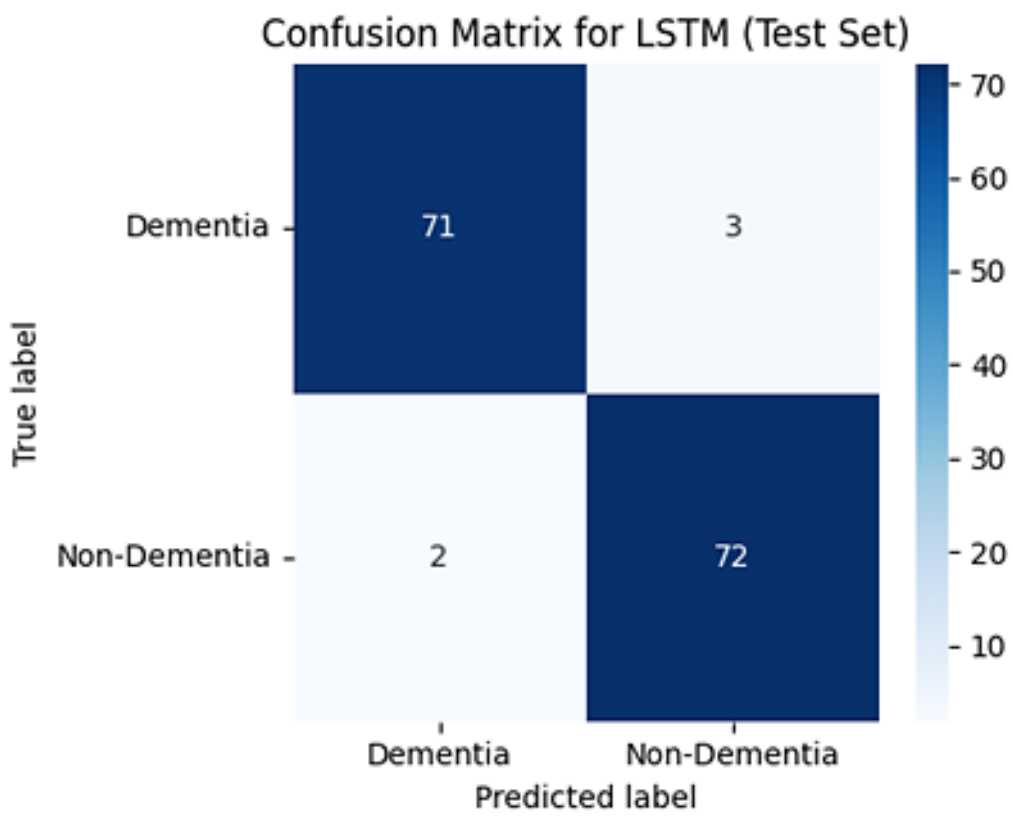

**Figure 7:** Confusion Matrix for Voice Classification Models Using an LSTM Network.

The confusion matrix (Figure 7) summarizes the LSTM model's performance in distinguishing dementia-related speech patterns from non-dementia cases. This model achieved 71 true positives and 72 true negatives, with only 5 misclassifications across 148 test samples. Gradient Boosting performs similarly to the LSTM network. Random Forest achieved 72 true positives and 71 true negatives, showing reliable accuracy. Logistic Regression struggles the most, with 3 false negatives and 10 false positives, making it less reliable. Overall, Gradient Boosting and LSTM are the best options for accurate dementia detection. But LSTM is a solid choice for dementia classification because of its sequential advantage. An important observation from the confusion matrix is the relatively small number of false negatives. In clinical screening systems, minimizing false negatives is particularly important because undetected dementia cases may delay medical intervention. The LSTM model's low false-negative rate suggests it may serve as a reliable tool for early dementia screening.

**Table 5:** Voice classification model’s performance.

| *Model* | *Accuray* | *Classes* | *Precision* | *Recall* | *F1-Score* |
|---|---|---|---|---|---|
| Logistic Regression | 91.22% | Dementia | 88.0 | 96.0 | 92.0 |
| | | Non-Dementia | 96.0 | 86.0 | 91.0 |
| Random Forest | 96.62% | Dementia | 96.0 | 97.0 | 97.0 |
| | | Non-Dementia | 97.0 | 96.0 | 97.0 |
| Gradient Boosting | 96.62% | Dementia | 97.0 | 96.0 | 97.0 |
| | | Non-Dementia | 96.0 | 97.0 | 97.0 |
| LSTM | 97.30% | Dementia | 97.0 | 96.0 | 97.0 |
| | | Non-Dementia | 96.0 | 97.0 | 97.0 |

Based on the models presented in Table 5, a comparative analysis reveals significant differences in their capabilities. Models like Random Forest and Gradient Boosting, although competitive with an accuracy of 96.62%, didn’t perform as well as LSTM. These models rely on decision trees that analyze individual features without considering the temporal relationships between them. While effective with structured data, they fall short in tasks like speech analysis, where the data sequence is critical. Similarly, Logistic Regression, a simpler linear model, achieved an accuracy of 89.86%. The method also struggled to accurately capture the subtle nuances of speech patterns and the gradual, unpredictable shifts within them over time. The LSTM model outperformed others, achieving an accuracy of 97.30%. This success is largely due to the unique features of the LSTM network, which excels at handling sequential and temporal data, especially for speech analysis and the identification of subtle changes that characterize speech patterns. In dementia research, speech patterns develop over time and often shift gradually. LSTM is capable of retaining information from previous steps and helps it to detect these changes. Such as slower speech and repetitive patterns, that might otherwise be difficult to notice in a single moment of speech. Throughout the training process, our designed LSTM model showed consistent improvement, with accuracy rising from 72.30% to 97.30% over 50 epochs. This consistent progress demonstrates LSTM's ability to learn complex patterns and adapt to nuanced changes in speech over time. Rather than analyzing isolated speech samples, LSTM learned the context and progression of speech, which is important for identifying progressive neurodegenerative conditions. The sequential advantage of LSTM is another reason for its success. While models like Random Forest and Gradient Boosting can perform well on numerous tasks, they tend to treat data as independent, static observations. This approach falls short when dealing with speech, where changes unfold gradually and over time. LSTM, however, processes speech as a time-dependent series, allowing it to capture the developing nature of speech patterns. This ability gives it a significant edge in identifying dementia, as early signs, such as slow speech or pauses, often emerge in ways that are not immediately apparent in a brief conversation.

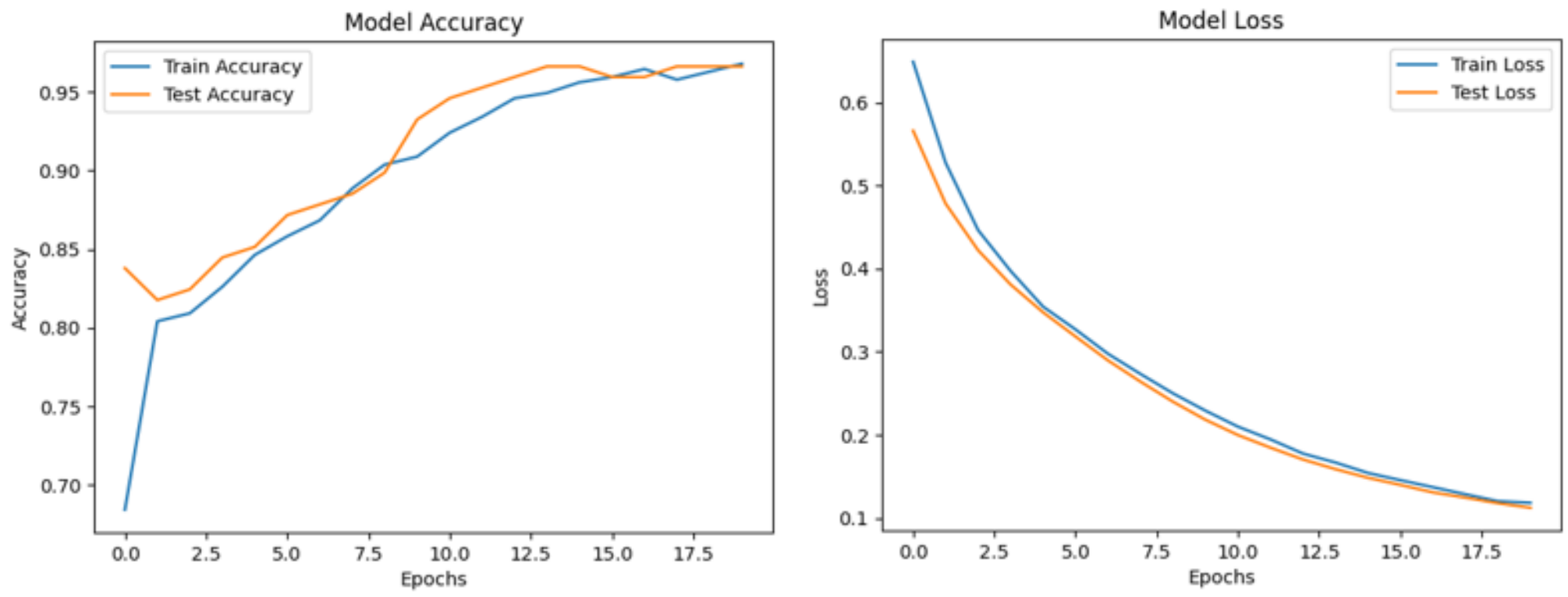

**Figure 8:** Performance of the LSTM model for voice classification.

Figure 8 shows two key performance metrics for a deep learning model: accuracy and loss over training epochs. The left plot shows the accuracy curves for the training and test sets, with a steady increase over epochs; the test accuracy slightly surpasses the training accuracy at times, highlighting good generalization. Early stopping is a regularization technique that halts training when the validation loss stops improving, thereby preventing overfitting. In our LSTM model, this technique slightly increased accuracy but prevented the model from reaching its maximum epoch of 50. This indicates that optimal performance was achieved in earlier training. The right plot shows the loss curves. Demonstrating a consistent decrease in both sets and signifying effective model convergence. In conclusion, our study shows that LSTM networks are exceptionally well-suited for detecting dementia through speech analysis compared to other machine learning models because they retain memory of previous time steps. By capturing dependencies and speech patterns, LSTM outperformed for early diagnosis. Neurodegenerative diseases often manifest as gradual, incremental changes in speech.

### *4-2- Performance Evaluation of Text-Based Classification*

The performance of the implemented models for dementia detection was evaluated using both acoustic analysis (voice data) and linguistic analysis (transcribed text data). While the voice classification achieved remarkable results with 740 interviewed voice samples, the transcribed text data posed challenges due to its limited size, particularly for models like LSTM that typically require large datasets. Despite this limitation, all models demonstrated competitive performance, with the LSTM model emerging as the most balanced and robust approach in all categories.

**Table 6:** Text classification model's performance.

| *Model* | *Accuracy* | *Classes* | *Precision* | *Recall* | *F1-score* |
|---|---|---|---|---|---|
| Logistic Regression | 83.7% | Dementia | 89.1 | 77.0 | 82.6 |
| | | Non-dementia | 79.8 | 90.5 | 84.8 |
| Random Forest | 72.9% | Dementia | 66.0 | 88.0 | 75.0 |
| | | Non-dementia | 82.0 | 54.0 | 65.0 |
| Support Vector Machine | 81.7% | Dementia | 85.1 | 77.0 | 80.9 |
| | | Non-dementia | 79.0 | 86.5 | 82.6 |
| LSTM | 82.4% | Dementia | 79.0 | 88.0 | 83.0 |
| | | Non-dementia | 86.0 | 77.0 | 81.0 |

The performance metrics presented in Table 6 indicate that the Logistic Regression (LR) model achieved a test accuracy of 83.7% and a cross-validation accuracy of 76% ±5%. The model performed well in classifying non-dementia cases, with a precision of 79.8% and a recall of 90.5%. However, for Dementia cases, recall was lower (77.0%), indicating that the model misclassified some cases. The F1-Scores for the dementia and non-dementia classes were 82.6% and 84.8%, respectively, indicating balanced performance. On the other hand, the Random Forest (RF) model achieved a test accuracy of 72.9% and a training accuracy of 100%, indicating significant overfitting. The model performed well at identifying dementia cases, with a high recall of 93.2%, but struggled with non-dementia cases, achieving a recall of only 52.7%. The imbalance is reflected in both the F1 Scores and the confusion matrix shown in Figure 9. The Support Vector Machine (SVM) model achieved a test accuracy of 81.7%, with a cross-validation accuracy of 76% ± 3%. This model also demonstrated balanced performance across both classes, with precision and recall of 85.1% and 77.0% for dementia, and 79.0% and 86.5% for non-dementia, respectively. The confusion matrix showed the model had a slight bias toward non-dementia classes. Lastly, the LSTM model achieved a test accuracy of 82.4%, with training and validation accuracies of 99.0% and 81.8%, respectively. Despite the limited size of the transcribed text dataset, the model demonstrated balanced precision and recall for both dementia (79.0%, 88.0%) and non-dementia (86.0%, 77.0%) classes. The confusion matrix in Figure 9 shows that the model misclassified 9 dementia cases and 17 non-dementia cases. Several misclassified cases, particularly among dementia samples, are also revealed by the confusion matrix. This may be attributed to the linguistic similarity between mild dementia speech and normal aging speech patterns.

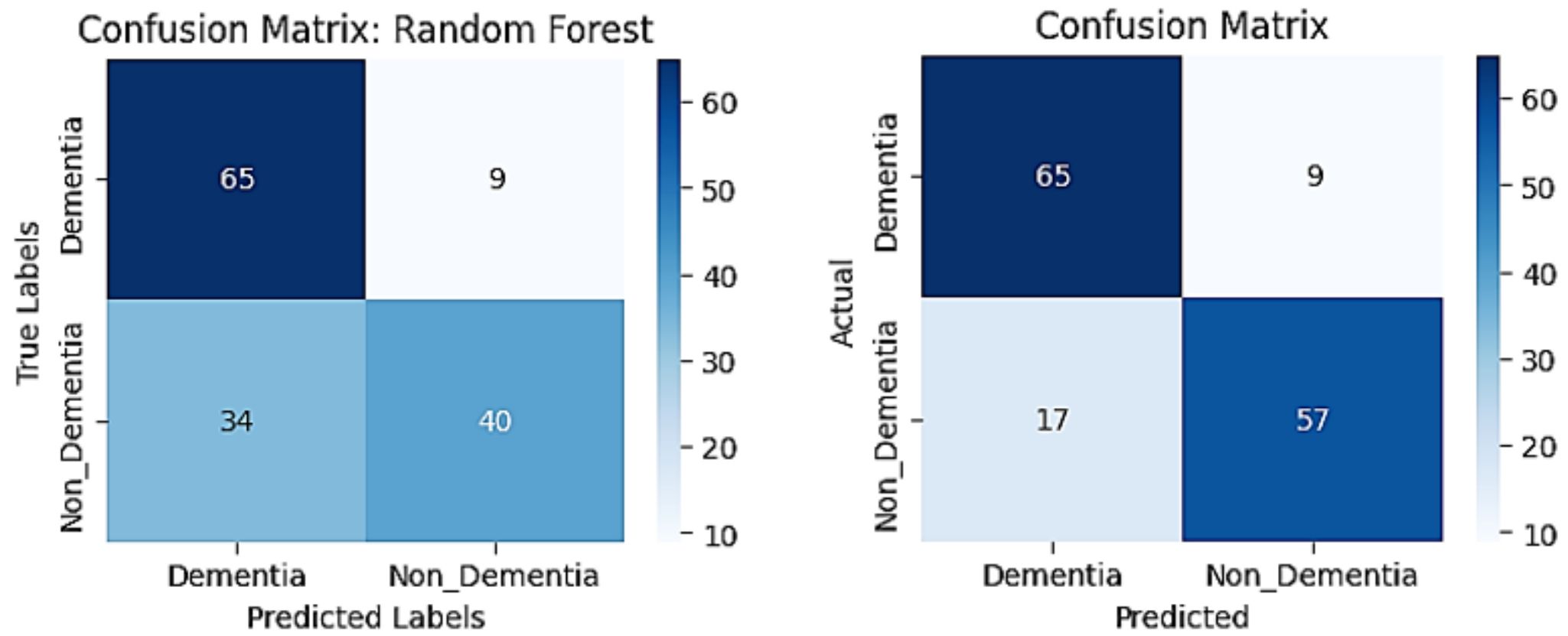

**Figure 9:** Confusion matrix for the least-performing RF and the best-performing LSTM model.

Compared with voice-based classification models, text-based models perform slightly worse. One possible reason is that transcription removes several important acoustic cues present in speech signals. Features such as speech rate, pause duration, pitch variation, and hesitation patterns often reflect cognitive decline in patients with dementia. These acoustic characteristics are not preserved in textual transcripts, limiting the ability of text-based models to fully capture dementia-related indicators. The Random Forest model exhibited clear signs of overfitting, as indicated by the large gap between training and testing accuracy. This behaviour suggests that the model memorized patterns specific to the training dataset rather than learning generalizable linguistic features. The relatively small size of the transcribed dataset likely contributed to this issue, as tree-based ensemble models can easily overfit when the sample size is limited. On the other hand, the performance of Logistic Regression suggests that certain linguistic indicators of dementia may be linearly separable within the feature space derived from the textual data. Features such as word frequency, sentence structure patterns, and lexical usage may provide meaningful signals that allow linear models to distinguish between dementia and non-dementia speech transcripts. Furthermore, the LSTM model demonstrates balanced precision and recall across both classes, highlighting its ability to capture sequential relationships within textual data. In natural language, the meaning of a sentence depends on the order and context of words. Dementia-related language impairments may manifest through unusual sentence structures, starting or ending with the same phrases, repeating words within a single sentence, or incomplete thoughts. By modelling these sequential dependencies, LSTM networks can identify subtle linguistic patterns that traditional machine learning algorithms may miss.

### *4-3- Performance Evaluation of the Sentence Auto-Completion Model*

Following the evaluation of text classification models, the sentence auto-completion task was addressed using a Bidirectional LSTM (Bi-LSTM) model. As mentioned in the methodology, the dataset, specifically designed for dementia patients, was divided into training, validation, and test sets to ensure robust evaluation. The Bi-LSTM model achieved training, validation, and test accuracies of 60.75%, 61.35%, and 59.74%, respectively, as shown in the training and validation accuracy curve in Figure 10. While these accuracy values may appear moderate, they are reasonable given the task's complexity and the dataset's limited size. The model's ability to predict the next word in a sequence was further evaluated using perplexity, a common metric for language models. The training perplexity was 5.27, the validation perplexity was 6.13, and the test perplexity was 6.52.

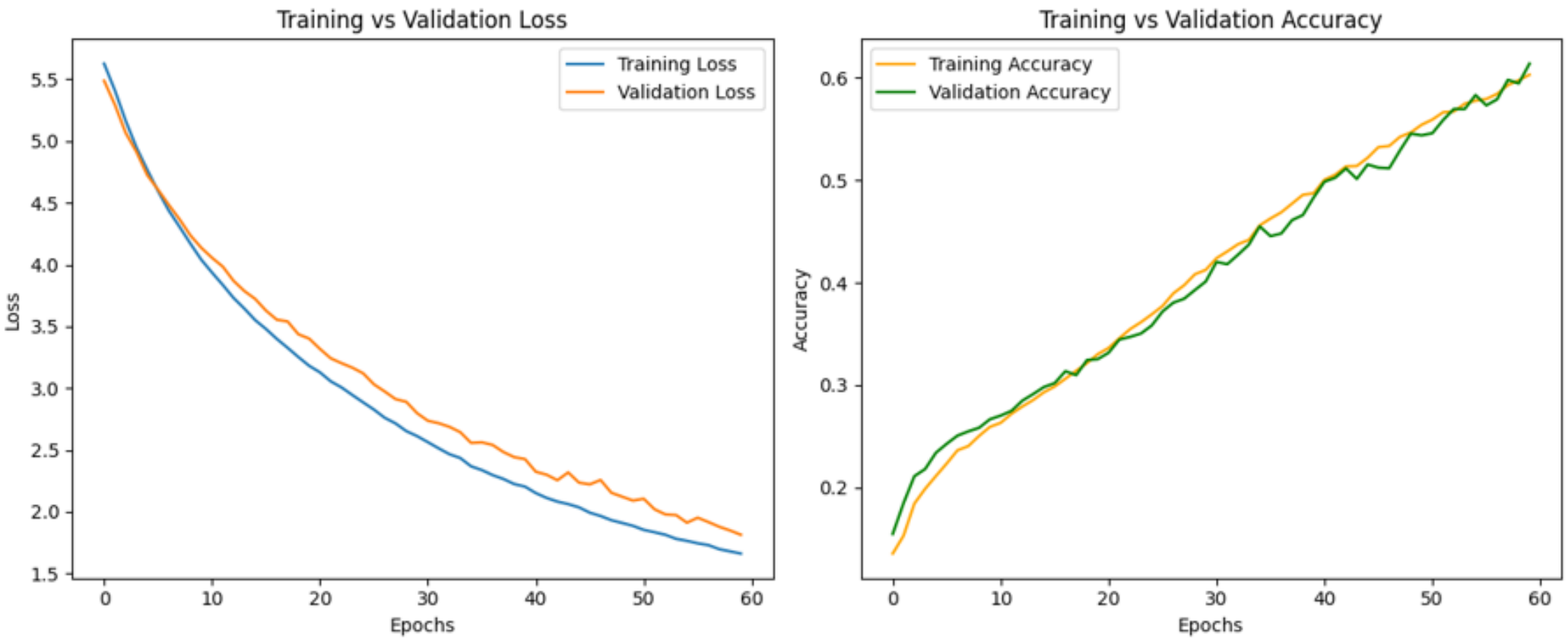

**Figure 10:** Accuracy and loss curve of the sentence completion Bi-LSTM model.

It is important to note that the actual performance of such models is often best evaluated through manual testing. During manual evaluation, the Bi-LSTM auto-completion model demonstrated outstanding performance on sentences similar to those in the training dataset, generating contextual and meaningful completions, as shown in the outputs in Table 7. However, in some cases, the model predicted meaningless or irrelevant words, leading to fluctuations in performance. This behavior is expected, given the limited size of the dataset and the inherent complexity of natural language generation. Despite this, the Bi-LSTM model showed promising results, achieving reasonable accuracy and low perplexity.

**Table 7:** Sentence completion evaluation.

| ***Input*** | ***Output*** |
|---|---|
| এবার করবানির ছাগলের …<br>*(This time, the sacrificial goat's …)* | এবার করবানির ছাগলের দাম অনেক বেশি দাও<br>*(This time, the sacrificial goat's price is very high)* |
| সকালে আমি নাস্তা করেছিলাম …<br>*(I had breakfast in the morning …)* | সকালে আমি নাস্তা করেছিলাম কিনা মনে পড়ছে না<br>*(I had breakfast in the morning, I can't remember)* |
| আমার রুটি দিয়ে হাঁসের …<br>*(with my flat rounded bread, to eat duck's)* | আমার রুটি দিয়ে হাঁসের মাংস খেতে ইচ্ছা করছে<br>*(with my flat rounded bread, to eat duck's meat, desire is happening)* |

## 5- Discussion

### *5-1- Transcription Alignment and Its Impact*

Accurate automatic speech recognition (ASR) for Bangla remains considerably more challenging than for high-resource languages such as English, particularly in speech affected by clinical conditions. In this study, we employed a transformer-based Bangla ASR pipeline to transcribe recorded speech into text. While the model performs robustly under standard conditions, transcription quality is inherently dependent on audio clarity and speaker articulation. In participants with dementia, speech production is often affected by hesitations, reduced fluency, articulation inconsistencies, and variable pronunciation quality, due to cognitive and neurological factors. As a result, we observed substantially higher transcription errors in dementia speech compared to non-dementia controls. To mitigate this issue and prevent errors in subsequent text-based analysis, we implemented a similarity-based sentence-matching approach using predefined prompts and a controlled similarity threshold. Quantitatively, the raw ASR output yielded an average similarity score of 86.74% with a word error rate (WER) of 0.46. After applying similarity-based sentence alignment, the average similarity increased to 98.10% and WER decreased to 0.10. These improvements indicate that the correction mechanism primarily addresses ASR-induced noise rather than altering semantic content.

It is important to clarify that increasing transcription accuracy was not the primary objective of this study. The goal was to ensure reliable textual input for subsequent analysis. Future work will focus on expanding the dataset and improving recording quality to evaluate system performance on more spontaneous, unseen speech without relying on sentence matching.

### 5-2- Cross-Study Performance Analysis

#### 5-2-1. Voice-Based Dementia Detection: Comparative Performance Analysis

Recent studies have investigated dementia detection through acoustic speech analysis across different linguistic settings. A 2025 study developed a Japanese conversational speech dataset collected from a memory clinic, comprising 263 training samples and a separate test set of 20 samples [31]. Using HuBERT, MFCC, F0, and silence-based features with a Bi-LSTM architecture, they reported 95% accuracy. Although the performance is promising, the extremely small test size, lack of cross-validation, and restriction to participants aged >=60 limit the statistical robustness of the findings. Similarly, Kumer et al. (2022) used 442 English recordings from the Pitt Corpus and extracted conventional acoustic features, including MFCC, log-energy, jitter, and shimmer [11]. Their best performance was 87% accuracy with Random Forest, while deep learning models (CNN, LSTM) achieved slightly lower results. However, the study remained confined to speech-only and did not explore multimodal modeling or assistive applications.

In contrast, the proposed study focuses on Bangla and addresses the scarcity of dementia-related speech resources for underrepresented linguistic groups. Our dataset is larger and class-balanced, improving statistical reliability and reducing classification bias. Using multiple machine learning and deep learning models, we achieved 97% accuracy with LSTM, 96% with RF and GB, and 91% with LR. Furthermore, beyond voice-based detection, our framework extends toward a border-assistive paradigm, which is based on text analysis and sentence completion.

#### 5-2-2. Text-Based Dementia Detection: Comparative Performance Analysis

Text-based dementia detection research has largely centered on the Pitt Corpus (Cookie Theft task) from DementiaBank, often under highly controlled experimental settings. A 2024 study introduced a topic-tracking LSTM framework that shows how subjects shift attention across predefined regions of the picture [32]. Using 552 transcribed speeches (309 from Alzheimer's disease patients, 243 from healthy controls), they achieved subject-level accuracy of 82.9%. However, the approach is intrinsically stimulus-dependent, as region-based modeling is a single picture description task, limiting its applicability to spontaneous speech. Likewise, Lin and Washington (2024) evaluated test-based classification using 9,447 sentence-level samples from the same dataset [33]. Their Word2Vec-LSTM model achieved approximately 60% accuracy without augmentation, rising to ~80% with synonym-based augmentation. Although augmentation improved performance, the reliance on a single benchmark dataset concerns external validity. Moreover, both studies primarily emphasize model performance optimization, with limited exploration of interpretable linguistic patterns that could provide clinically meaningful insights.

In contrast, the proposed study is based on a newly collected Bangla-language dataset rather than a pre-existing corpus. All speeches were independently recorded and transcribed through a transformer-based whisper model. Unlike prior studies that depend on task-specific or artificial text augmentation, our approach evaluates models on authentic, non-augmented data to avoid potential distributional distortion and overestimation of performance. Using TF-IDF representations with LR, RF, SVM, and LSTM classifiers, we achieved 83.7% accuracy with LR and 82.4% accuracy with LSTM. Comparable performance without synthetic augmentation supports the robustness of the proposed framework and strengthens its generalizability beyond stimulus-specific experimental settings.

#### 5-2-3. Sentence Completion Approaches: Comparative Performance Analysis

Next-word prediction, or sentence completion, has been widely studied in high-resource, general-domain language tasks. However, its application for dementia-oriented communication assistance remains largely unexplored. For example, Biswas et al. (2026) investigated next-word prediction in Bengali poetry using a corpus of 6,706 poems from 137 poets. They evaluated LSTM, BiLSTM, Transformer Encoder, and fine-tuned BanglaBERT models, achieving a maximum validation accuracy of 0.77 with the Transformer Encoder [34]. Similarly, a 2024 study on Urdu developed LSTM- and BERT-based next-word prediction models using approximately 1.1 million sentences [35]. The author reported 52.4% accuracy with LSTM and 73.7% with BERT-based models. The large-scale corpus made transformer-based architecture feasible and allowed extensive pretraining and fine-tuning. Nevertheless, the study focused on general language modeling and did not address cognitively constrained speech patterns or assist in communication.

In contrast, the proposed study introduces a novel application domain: sentence auto-completion specifically designed to support communication for dementia patients in Bangla. To the best of our knowledge, no prior study has developed a prediction architecture explicitly targeting dementia-related conversational assistance. Unlike the aforementioned works, our dataset was newly collected from a clinically relevant population and is substantially smaller due to the inherent challenges of recruiting and transcribing speech data from people with dementia. Consequently, large transformer-based or pretrained language models were not feasible without risking overfitting or distributional mismatch. Despite these constraints, our systematic modelling approach using a Bi-LSTM architecture achieved a validation accuracy of 61.35% with a perplexity of 6.13. Considering the limited dataset size, the results demonstrate that meaningful contextual prediction is achievable under clinically realistic data constraints.

## 6- Conclusion & Future Work

This study addresses critical gaps in dementia detection and communication assistance for the Bangla-speaking population by developing a comprehensive multimodal framework using voice and text data. The proposed system integrates acoustic feature analysis and linguistic modeling to support classification and conversational assistance with advanced machine learning and deep learning techniques. A custom dataset of Bangla voice recordings and sentences was developed to reflect daily communication needs. Experimental results demonstrated that deep learning models are effective for dementia-related tasks, with the LSTM model achieving 97.03% accuracy in voice-based classification and 82.4% in text-based classification. Furthermore, the Bidirectional LSTM (Bi-LSTM) model with an attention mechanism for sentence auto-completion achieved a test accuracy of 59.74%, demonstrating its potential to assist patients with cognitive impairment in real-time communication. The results indicate that combining acoustic and linguistic features provides a reliable approach for dementia analysis in an underrepresented language.

The proposed framework contributes to dementia research by introducing a Bangla-specific multimodal dataset and an integrated system for both detection and communication support. Despite the promising results, the study has several limitations, including a relatively small dataset and the use of controlled reading tasks rather than spontaneous speech. Future work will focus on expanding the dataset to include more diverse participants and on incorporating more impulsive conversational speech to improve real-world applicability. Further improvements may be achieved by adopting advanced model architectures, such as transformer-based language models. Increasing the dataset size would enable the models to learn more dynamic voice patterns and a wider range of sentence structures. Although collecting and processing Bangla data was challenging, the proposed approach demonstrates strong potential for supporting dementia patients. The developed framework can also be extended to other underrepresented languages in future research.

## 7- References